%% file: main.tex
\documentclass[11pt]{article}

\usepackage[preprint]{acl}

\usepackage{times}
\usepackage{latexsym}

\usepackage[T1]{fontenc}

\usepackage[utf8]{inputenc}
\usepackage{tikz}
\usepackage{amsmath}
\usetikzlibrary{shapes,arrows,positioning,fit,backgrounds}
\usepackage{microtype}

\usepackage{inconsolata}

\usepackage{graphicx}
\usepackage{tikz}
\usepackage{amsmath}
\usepackage{amssymb}
\usetikzlibrary{shapes,arrows,positioning,fit,backgrounds}

\usepackage{booktabs}
\usepackage{makecell}

\title{Transition-Matrix Regularization for Next Dialogue Act Prediction in Counselling Conversations}

\author{
Eric Rudolph \and Philipp Steigerwald \and Jens Albrecht \\
Technische Hochschule Nürnberg Georg Simon Ohm \\
\texttt{\{eric.rudolph, philipp.steigerwald, jens.albrecht\}@th-nuernberg.de}
}
\begin{document}
\maketitle
\begin{abstract}
This paper studies how empirical dialogue-flow statistics can be incorporated into Next Dialogue Act Prediction (NDAP). A KL regularization term is proposed that aligns predicted act distributions with corpus-derived transition patterns. Evaluated on a 60-class German counselling taxonomy using 5-fold cross-validation, this improves macro-F1 by 9--42\% relative depending on encoder and substantially improves dialogue-flow alignment. Cross-dataset validation on HOPE suggests that improvements transfer across languages and counselling domains. In systematic ablations across pretrained encoders and architectures, the findings indicate that transition regularization provides consistent gains and disproportionately benefits weaker baseline models. The results suggest that lightweight discourse-flow priors complement pretrained encoders, especially in fine-grained, data-sparse dialogue tasks.
\end{abstract}

\section{Introduction}

Next dialogue act prediction (NDAP) forecasts the communicative function of the \emph{upcoming} utterance from the dialogue history. Although the task has a long tradition in dialogue research \citep{nagata_first_1994, stolcke_dialogue_2000, reithinger_predicting_1996}, it has received less attention in the era of large language models (LLMs). Yet it offers a structured and interpretable mechanism for steering LLM behavior, complementing prompting \citep{brownLanguageModelsAre2020}, instruction tuning \citep{wei2022finetuned}, and reward-based approaches \citep{ouyang_training_2022}. By anticipating the next dialogue action, systems can condition prompts or constraints to encourage more stable, coherent, and goal-directed behavior \citep{chen_controllable_2023}.

\input{figures/contribution_figure}

In counselling and other highly structured domains, next acts follow consistent pragmatic patterns: greetings typically precede problem statements, exploration precedes intervention, and closing behaviors follow resolution \citep{bickmore_automated_2013, althoff_large-scale_2016}. Classical dialogue managers explicitly modeled these transitions through Markov or CRF-based structures \citep{stolcke_dialogue_2000, zimmermann_joint_2009}. Modern neural systems, however, largely abandon symbolic transition models and instead rely on end-to-end architectures to infer discourse structure implicitly \citep{ultes_pydial_2017, ravuru_multi-domain_2022}.

This shift removes an inductive bias. Neural models see only a single gold next-act label per instance, providing limited signal when several next acts are plausible—a common situation in counselling \citep{wu_towards_2022, demasi_multi-persona_2020}. The gold label in NDAP is inherently under-specified: it represents one observed continuation among many valid possibilities. Standard cross-entropy supervision thus penalizes the model for predicting other plausible acts. Consequently, models may struggle to capture the distribution over multiple valid next actions \citep{zhao_learning_2017}.

This limitation is addressed by incorporating an empirical transition matrix directly into the loss. The \textbf{transition-matrix KL regularizer} encourages the predicted next-act distribution to align with observed transition statistics, injecting pragmatic discourse-flow information as a soft, differentiable constraint (Figure~\ref{fig:contribution}). This preserves the flexibility of neural encoders while reinstating a structural prior reminiscent of classical systems.

This idea is evaluated in German text-based counselling, where communicative actions are fine-grained and governed by psychosocial norms. The dataset uses a five-level taxonomy with 60 dialogue act (DA) categories \citep{albrecht_oncoco_2025}. NDAP is performed across all speaker transitions. To exploit the taxonomy structure, category history augmented architectures are introduced. 

The results show that transition-based regularization provides consistent gains and disproportionately benefits weaker models. The regularizer is lightweight, model- and architecture-agnostic, and can be integrated as a drop-in objective without architectural modifications or sequence-level decoding. Improvements span both predictive metrics (F1, Top-3) and structural measures of dialogue-flow alignment.

\section{Related Work}

\subsection{NDAP and Future-Act Prediction}

Dialogue act (DA) prediction is a well-established task that assigns communicative functions to observed utterances. Classical systems model discourse structure using stochastic grammars, HMMs, or CRFs \citep{stolcke_dialogue_2000, geertzen2009dialogue}, while recent neural approaches employ hierarchical encoders, contextual attention, and multimodal cues \citep{colombo_guiding_2020}. These models capture local and long-range structure but operate entirely on \emph{observed} utterances.

In contrast, our work addresses the harder task of NDAP: forecasting the communicative function of the \emph{upcoming} utterance without access to its surface form. Early statistical work showed that DA sequences exhibit strong structural regularities, with n-gram models improving prediction through explicit transition constraints \citep{reithinger_predicting_1996, geertzen2009dialogue}. Neural NDAP research remains limited. Prior work integrates multi-turn history using hierarchical attention \citep{tanaka2019neural}, incorporates multimodal features, or employs semi-supervised consistency objectives \citep{he2022galaxy}. Latent-variable architectures \citep{ji2016latent} regularize discourse trajectories, while future-act prediction has also been studied in adjacent settings such as classroom talk moves and counselling response-act forecasting \citep{ganesh_what_2021, wu_towards_2022, srivastava_response-act_2023}. Overall, existing NDAP-style approaches condition on DA history but do not constrain predictions using empirical dialogue-flow statistics.

\subsection{Explicit Transition Constraints and Posterior Regularization}

Explicit modeling of DA transitions is well explored in \emph{sequence labeling}, where the goal is to assign a DA label to every utterance in a conversation. Classical systems encode transitions through n-gram dialogue grammars or HMMs \citep{stolcke_dialogue_2000}, and neural architectures commonly add CRF layers to impose label-transition structure \citep{chenDialogueActRecognition2018, shang_speaker-change_2020}. These methods learn transition potentials or decode full DA sequences with structural constraints. However, they do not address NDAP, where only the \emph{next} act must be predicted and no sequence decoder is used.

Distribution-level constraints have been introduced through posterior regularization, which biases models toward constraint-satisfying priors using KL divergence \citep{JMLR:v11:ganchev10a}. Such techniques have improved dialogue understanding and state tracking \citep{jin_explicit_2018}, and future-aware constraints have been applied to generation models \citep{feng_regularizing_2020}. These works demonstrate the utility of KL-based structural guidance, but they do not incorporate corpus-derived DA-transition statistics directly into the NDAP objective.

\subsection{LLM-Based Dialogue Control and Planning}

Recent approaches increasingly rely on LLMs for dialogue generation, typically steered via prompting, control signals, latent dialogue actions, or search-based planning. Examples include controller-guided generation \citep{shukuri_meta-control_2023, wagner_controllability_2024}, latent dialogue-act control \citep{wu_diacttod_2023}, structure-aware task-flow modeling \citep{sohn_tod-flow_2023}, and prompt-based policy planning with Monte Carlo Tree Search \citep{yu_prompt-based_2023}. In counselling, LLM-based systems have been explored for response generation, counselor-facing support, and virtual client simulation \citep{srivastava_response-act_2023, steigerwald_caia_2025, rudolph_comparing_2025}.

However, growing evidence suggests that LLMs do not reliably acquire human-like discourse behavior, particularly with respect to pragmatic sequencing and role consistency \citep{shukuri_meta-control_2023, wagner_controllability_2024}. Fluent surface realization therefore provides limited leverage for structured dialogue control, motivating the use of explicit structural priors.

Our approach fills a gap between these research threads. Unlike NDAP models that rely on implicitly learned transitions, we introduce an explicit, data-derived \emph{transition-matrix regularizer} that aligns the predicted next-act distribution with empirical dialogue-flow patterns. Unlike CRF or HMM models, our method does not require sequence decoding. And unlike posterior-regularization approaches, our structural prior is grounded directly in observed DA transitions. To our knowledge, this is the first instance of integrating empirical DA-transition constraints into the optimization objective for NDAP.

\section{Method}

\subsection{Problem Formulation}

Given a conversation history consisting of $n$ utterances $\{u_1, u_2, \ldots, u_n\}$, the goal is to predict the category $c_t$ of the next utterance. All speaker transitions are considered (Counselor$\to$Counselor, Counselor$\to$Client, Client$\to$Client, Client$\to$Counselor), performing NDAP based on the conversation history. Categories belong to a five-level hierarchy with 60 leaf-level dialogue acts.

The task assumes access to gold DA labels in conversation history, appropriate for post-hoc analysis, training simulations, and constrained LLM generation.

To ground the task in a concrete taxonomy, the full hierarchical structure is adopted from the OnCoCo dataset, which organizes 60 leaf-level categories into progressively more abstract semantic groups \citep{albrecht_oncoco_2025}. The hierarchy captures both conversational function and pragmatic counseling flow: high-level groups distinguish phases such as greetings, problem exploration, motivation building, and closing, while lower levels specify fine-grained communicative functions such as factual problem descriptions, emotional disclosures, resource activation, or evaluation of solution attempts.

OnCoCo provides finer granularity (60 categories vs.\ 3--15 in alternatives such as Anno-MI \cite{wu_anno-mi_2022}, HOPE \cite{malhotra_speaker_2022}, or MITI \cite{moyers_motivational_2016}) and integrates multiple counselling paradigms rather than a single therapeutic approach, critical for controllable client simulation.

\subsection{Architectures}

\input{history_model_fig}

The base model encodes utterances via a pretrained BERT encoder and aggregates token representations through learned attention pooling. The pooled representation is passed to a linear classifier for NDAP.

The enhanced variant replaces attention pooling with multi-head self-attention for utterance encoding and adds 8-head cross-utterance attention to model conversation flow. A sliding window of the previous $h$ categories is embedded with positional encodings and processed through 8-head self-attention. A text-category cross-attention mechanism integrates the historical context with the conversation representation, and a 4-head category transition attention layer models dependencies between consecutive acts (Figure~\ref{fig:history_model}).

\subsection{Transition Matrix and Dialogue Flow Regularization}

A central component in our approach is the \textit{transition matrix loss}, which encourages predictions to respect observed category transition patterns in conversation. Although prior dialogue-act models have encoded transitions implicitly through sequential architectures or explicitly through statistical models, to the authors' knowledge no prior work incorporates an empirical transition matrix directly into the training objective of a neural NDAP model.

\subsubsection{Computing the Empirical Transition Matrix}

\begin{figure}[t]
  \centering
  \includegraphics[width=0.8\columnwidth]{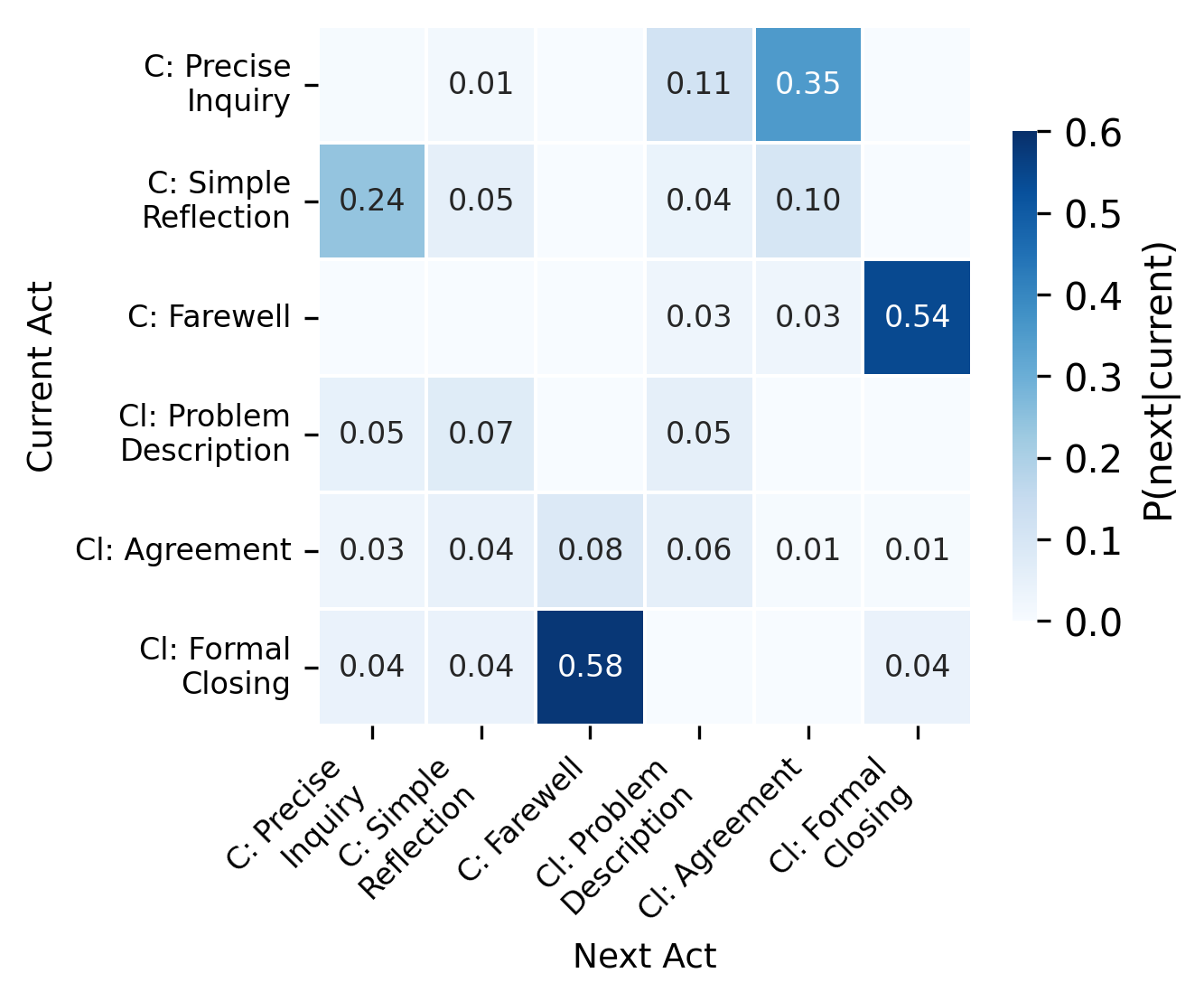}
  \caption{Example 6x6 subset (Fold 0) of the empirical transition matrix. C = Counselor, Cl = Client. High probabilities show pragmatic patterns: closings trigger closings, precise inquiry lead to agreement.}
  \label{fig:transition_matrix}
\end{figure}

A transition matrix encodes the probability distribution over next states given the current state---here, the likelihood of each dialogue act following another. From the training corpus, we compute a normalized empirical transition matrix $\mathbf{T}$ where $T_{ij}$ represents the probability of transitioning from category $i$ to category $j$. To avoid zero probabilities in later KL computations, we add a fixed constant $\epsilon_{\text{num}} = 10^{-8}$ to all cells after row normalization; this value is used solely for numerical stability and is not treated as a modeling hyperparameter. In social counselling conversations, the transition matrix exhibits clear pragmatic patterns as can be seen in figure \ref{fig:transition_matrix}.

\subsubsection{Transition Matrix Loss}
\label{sec:tm_loss}

The transition matrix loss uses KL divergence to measure alignment between model predictions and empirical category transitions. Given the previous category $c_{t-1}$, the target transition distribution is:

\begin{equation}
  \mathbf{t}^{(t)} = \mathbf{T}[c_{t-1}] \in \mathbb{R}^C
\end{equation}

The model's predicted distribution is:

\begin{equation}
  \hat{\mathbf{p}}^{(t)} = \text{softmax}(\hat{\mathbf{y}}_{\text{final}})
\end{equation}

The transition matrix loss measures divergence:

\begin{equation}
  \mathcal{L}_{\text{TM}} = \text{KL}(\hat{\mathbf{p}}^{(t)} \| \mathbf{t}^{(t)}) = \sum_{j=1}^{C} \hat{p}_j^{(t)} \log \frac{\hat{p}_j^{(t)}}{t_j^{(t)}}
\end{equation}

KL divergence is asymmetric, penalizing impossible transitions more heavily than missing low-probability valid ones, making it well-suited for enforcing dialogue structure.

Our transition-matrix regularizer differs from label smoothing \citep{szegedy_rethinking_2015}, which distributes probability mass uniformly across non-target classes. In contrast, our approach distributes mass according to empirically observed transition patterns, providing domain-specific rather than uniform smoothing (Table~\ref{tab:ablation_ls_full}; see Appendix~\ref{sec:ablation} for additional comparisons on the client simulation subset).

The weight $\lambda_{\text{tm}} \in \{0.0, 0.2, 0.5, 1.0, 1.5\}$ is explored systematically (Section~\ref{sec:results}).

\subsection{Training Objective}

Models combine cross-entropy loss with transition-matrix regularization:
\begin{equation}
  \mathcal{L}_{\text{total}} = \mathcal{L}_{\text{CE}} + \lambda_{\text{tm}} \mathcal{L}_{\text{TM}}.
\end{equation}

Cross-entropy provides the primary supervision signal based on the single annotated next category, while $\mathcal{L}_{\text{TM}}$ (defined in Section~\ref{sec:tm_loss}) introduces a soft prior reflecting the empirical distribution over multiple valid next acts.

\section{Experiments}

\subsection{Dataset}

Models are evaluated on a corpus of German social counselling dialogues consisting of 76 conversations with 5,457 utterances. The data originate from structured counselling role-play sessions conducted by social science students in a university course on online text-based counselling. Participants acted in predefined client and counselor roles, and no real clients or personal information were involved. All conversations were collected with consent for research use.

\begin{table}[t]
  \centering
  \small
  \setlength{\tabcolsep}{3pt}
  \begin{tabular}{@{}>{\raggedright\arraybackslash}p{0.18\columnwidth}>{\raggedright\arraybackslash}p{0.25\columnwidth}>{\raggedright\arraybackslash}p{0.49\columnwidth}@{}}
    \toprule
    \textbf{Speaker} & \textbf{Category} & \textbf{Utterance} \\
    \midrule
    Client & Personal disclosure & ``Hello.'' \\
    Client & Problem description & ``My child is taking drugs. Can you help me here?'' \\
    Counselor & Opening & ``Hello. I am one of the counselors.'' \\
    Counselor & Inquiry about concern & ``How can I help you with this?'' \\
    Client & Problem definition & ``Because of the drugs, he is now having problems at school, and I do not want that for him.'' \\
    \bottomrule
  \end{tabular}
  \caption{Excerpt from an OnCoCo role-play conversation, translated from German for readability.}
  \label{tab:conversation_example}
\end{table}

Table~\ref{tab:conversation_example} illustrates the style of the role-play data and the granularity of the annotation scheme. Each utterance is annotated with a category from the OnCoCo taxonomy \citep{albrecht_oncoco_2025}, which defines 60 leaf-level dialogue act categories organized in a five-level hierarchy. While the taxonomy has been previously described, this work contributes the first publicly available corpus of complete counselling conversations annotated with this scheme, enabling sequential modeling of dialogue flow.

Annotation followed a two-stage workflow. Paid social science students with prior training in counselling concepts first labeled utterances with OnCoCo categories. A domain expert then reviewed each conversation sequentially, i.e., in dialogue order rather than as isolated utterances, and corrected labels where necessary. This protocol yields expert-reviewed annotations, but it does not produce a standard inter-rater agreement (IAA) statistic because conversations were not independently double-annotated. We therefore position the corpus as expert-reviewed rather than IAA-validated, and treat this as an explicit limitation when interpreting annotation reliability and downstream generalization.

The human-annotated conversational dataset, along with metadata describing the annotation schema, is released to support reproducibility and further research on hierarchical dialogue-act prediction and counselling dialogue modeling.\footnote{Code and data available at \url{https://github.com/rudolpheric/tm-reg}}

For experimentation, NDAP is performed across all speaker transitions, yielding instances across 60 categories. Evaluation uses 5-fold cross-validation with conversation-level partitioning to prevent data leakage. For each fold, training is conducted on 80\% of conversations with evaluation on the held-out 20\%. Mean performance $\pm$ standard deviation across all five test folds is reported. Rather than selecting optimal hyperparameters on a separate validation set (which would further reduce already limited training data), results across the full hyperparameter grid are reported (Section~\ref{sec:results}), allowing readers to assess performance at each regularization strength. The original class distribution exhibits substantial imbalance. LLM-based synthetic data augmentation was also explored to address this imbalance; however, it did not improve BERT-based models (see Appendix~\ref{sec:ablation_synth} for details).

\subsection{Baselines and Models}

The history-augmented architecture is compared against several baseline approaches: (1) a transition-matrix baseline that uses only empirical dialogue-flow patterns without learning, (2) a Simple RNN baseline, (3) the architecture proposed by \citet{tanaka2019neural}, which uses hierarchical attention to integrate multi-turn dialogue history for NDAP, and (4) zero-shot LLM baselines.

To validate the robustness of the findings across different pretrained language models, all neural baselines and history-aware models were tested using 7 different German BERT variants: EuroBERT-210m, EuroBERT-610m, G-BERT-base, G-BERT-large, GELECTRa-base, Modern-G-BERT-134M, and Modern-G-BERT-1B. Additionally, context window size is varied (1, 4, 8, 12 utterances) and transition-matrix regularization is compared against label smoothing ($\epsilon \in \{0.0, 0.1, 0.2\}$) as an alternative regularization strategy (see Appendix~\ref{sec:ablation}). This systematic evaluation across 300 configurations (7 encoders $\times$ 2 architectures $\times$ 5 TM weights $\times$ 4 context lengths, plus label smoothing variants), evaluated using 5-fold cross-validation, shows that architectural improvements hold consistently across language models ranging from 110M to 1B parameters. Table~\ref{tab:models} summarizes the model architectures and their properties.

\textbf{Transition Matrix Baseline:} The transition matrix baseline provides a competitive non-learning baseline. This model predicts the most likely next category given only the previous utterance's category, using the empirical transition matrix computed from training data. It requires no neural training or context encoding and serves as a practical reference point for the value of learned contextual representations.

\textbf{RNN Baselines:} Two RNN-based baselines are implemented: (1) Simple RNN, a basic recurrent architecture over utterance embeddings, and (2) the \citet{tanaka2019neural} architecture, which uses hierarchical attention mechanisms to model multi-turn context for NDAP. All RNN baselines are trained with the same transition-matrix regularization as BERT models. Results are shown in Section~\ref{sec:results_baseline}.

\textbf{LLM Baseline:} To contextualize fine-tuned BERT performance against state-of-the-art language models, both proprietary (GPT-5-mini) and open-source (gpt-oss-120b, a 120B-parameter Mixture of Experts model) LLMs are evaluated in a zero-shot setting. Each model receives conversation history (last 12 turns) and all 60 category descriptions, returning top-3 predictions with confidence scores. This baseline tests whether large-scale pretraining alone captures dialogue-flow patterns without explicit structural constraints. The full prompt template is provided in Appendix~\ref{sec:llm_prompt}.

\begin{table}[t]
  \centering
  \resizebox{0.7\columnwidth}{!}{
  \begin{tabular}{lcc}
    \toprule
    \textbf{Model} & \textbf{Attention} & \textbf{History} \\
    \midrule
    Transition Matrix & \textemdash & \textemdash \\
    GPT-5-mini (LLM) & \textemdash & \checkmark \\
    gpt-oss-120b (LLM) & \textemdash & \checkmark \\
    Simple RNN & Pooling & \textemdash \\
    Tanaka (2019) & Hier. Attn. & \checkmark \\
    BERT & Pooling & \textemdash \\
    BERT+History & Multi-head & \checkmark \\
    \bottomrule
  \end{tabular}
  }
  \caption{Model architectures and properties.}
  \label{tab:models}
\end{table}

\subsection{Evaluation Metrics}
\label{sec:metrics}

Models are evaluated on predictive correctness and dialogue-flow alignment via three metric categories:

\paragraph{Predictive Correctness} Macro-F1, weighted F1, and Top-3 accuracy serve as the primary measures. Top-3 accuracy acknowledges that multiple next acts can be plausible. Note that these metrics assume a single correct answer, which is a simplification: in NDAP, multiple dialogue acts are often genuinely valid continuations, so moderate absolute scores are expected.

\paragraph{Dialogue-Flow Alignment} To assess how well model predictions adhere to the conversational structure observed in the training data, a group of metrics based on a first-order empirical transition matrix $\mathbf{T}$ is employed: Cumulative Accuracy at 70\% (Cum70) and Jensen-Shannon (JS) Divergence. Cum70 measures whether the predicted category falls within the set of most likely transitions that together account for 70\% of the empirical probability mass---capturing whether predictions align with pragmatically plausible continuations. JS divergence measures the overall distributional alignment between predicted and empirical transition distributions.

\paragraph{Rationale for First-Order Metrics} While dialogue context is inherently rich and multi-turn, the dialogue-flow metrics deliberately rely on a first-order (last-act-to-next-act) transition matrix. This is a practical necessity. Constructing higher-order transition matrices (e.g., second-order, based on the last two acts) would be infeasible given the 60 leaf categories, as it would lead to a combinatorial explosion of states ($60^2 = 3600$ potential source pairs) and result in an extremely sparse and unreliable matrix with this dataset size. The first-order matrix thus serves as a robust and computable proxy for measuring the model's grasp of foundational, short-term dialogue coherence.

\paragraph{Transition Matrix Computation} The transition matrix is computed from training data within each CV fold, measuring whether the model internalized valid dialogue flows.

\begin{figure*}[t]
  \centering
  \includegraphics[width=\textwidth]{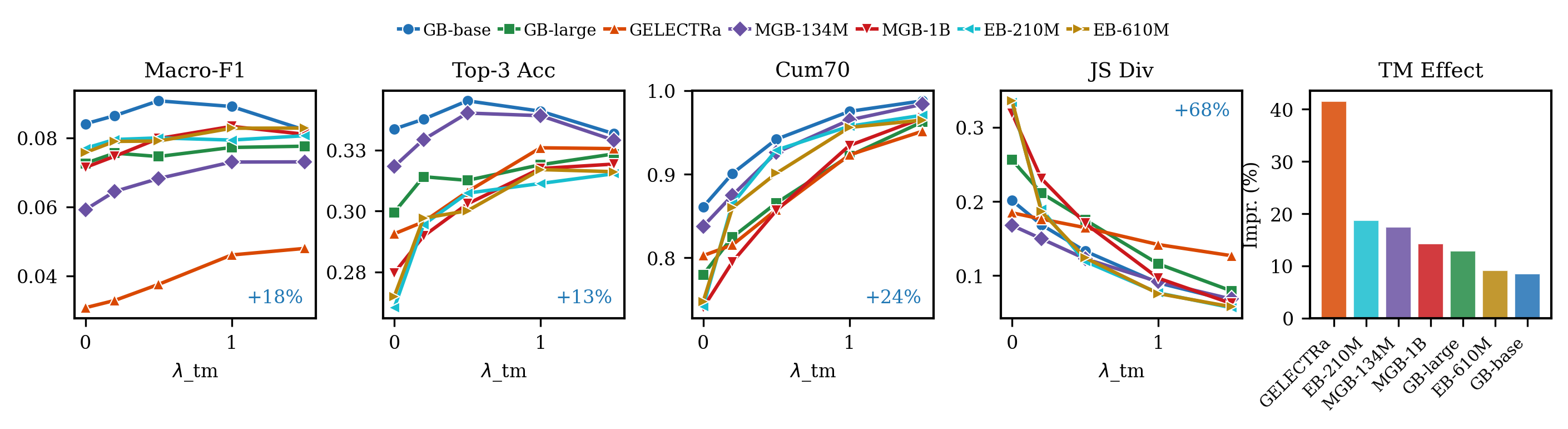}
  \caption{Effect of transition loss weight ($\lambda_{\text{tm}}$) on 60-category classification. Left four panels show how Macro-F1, Top-3 Accuracy, Cum70, and JS Divergence change with increasing $\lambda_{\text{tm}}$ across encoders. Right panel compares relative macro-F1 gains from TM regularization by encoder.}
  \label{fig:tm_effect}
\end{figure*}

\begin{table}[t]
  \footnotesize
  \setlength{\tabcolsep}{2pt}
  \resizebox{\columnwidth}{!}{
  \begin{tabular}{@{}lccccc@{}}
    \toprule
    \textbf{Encoder} & \textbf{Best$_{\lambda=0}$} & \textbf{Best$_{\lambda>0}$} & $\lambda$ & $\Delta$ & \textbf{\%Gain} \\
    \midrule
    GBERT-large & .097 & \textbf{.102} & 0.5 & +.005 & +5.1\% \\
    GBERT-base & .092 & .098 & 0.5 & +.005 & +6.0\% \\
    ModernGBERT-1B & .089 & .096 & 0.2 & +.007 & +8.5\% \\
    EuroBERT-610M & .087 & .096 & 0.5 & +.009 & +10.5\% \\
    EuroBERT-210M & .080 & .095 & 0.5 & +.014 & +18.0\% \\
    ModernGBERT-134M & .076 & .086 & 0.5 & +.009 & +12.2\% \\
    GELECTRa-base & .060 & .070 & 1.0 & +.009 & +16.4\% \\
    \midrule
    \textit{Mean} & .083 & .092 & -- & +.008 & +11.0\% \\
    \bottomrule
  \end{tabular}
  }
  \caption{TM regularization effect by encoder (macro-F1, 60 categories, 5-fold CV). Compares best result at $\lambda_{\text{tm}}$=0 vs best at $\lambda_{\text{tm}}$>0. Sorted by best macro-F1.}
  \label{tab:tm_effect}
\end{table}

\subsection{Results}
\label{sec:results}
\subsubsection{TM Regularization Effect by Encoder}
\label{sec:results_baseline}

Table~\ref{tab:tm_effect} shows the effect of transition-matrix regularization on the 60-category classification task, comparing each encoder's best result at $\lambda_{\text{tm}}$=0 versus its best result at $\lambda_{\text{tm}}$>0. 

All encoders improve with transition-loss regularization (Table~\ref{tab:tm_effect}). Relative gains range from +5.1\% (GBERT-large) to +18.0\% (EuroBERT-210M), with a mean improvement of +11.0\%. Notably, smaller encoders show larger relative gains, suggesting TM regularization partially compensates for limited model capacity. Optimal $\lambda_{\text{tm}}$ values cluster around 0.5, with only GELECTRa-base benefiting from higher regularization ($\lambda_{\text{tm}}$=1.0). Architecture effects vary by encoder (Appendix \ref{app:significance} Table~\ref{tab:arch_effect_by_encoder}).
Figure~\ref{fig:tm_effect} visualizes values averaged across architectures and context sizes for each $\lambda_{\text{tm}}$ value; this averaging reveals larger relative improvements, ranging from 9\% (GBERT-large) to 42\% (GELECTRa-base), consistent with the abstract.
\subsubsection{Cross-Dataset Validation}
\label{sec:cross_dataset}

To validate generalization beyond German counselling, the transition-matrix regularizer is evaluated on the HOPE dataset \cite{malhotra_speaker_2022}, an English counselling corpus with 15 dialogue act categories. Table~\ref{tab:cross_dataset} summarizes results using XLM-RoBERTa-base and BERT-base encoders with both BERT and History architectures. Additional evaluation on Switchboard (SWDA), a non-counselling benchmark with highly skewed transition distributions, is provided in Section~\ref{sec:ablation_kl}.

\begin{table}[t]
  \centering
  \footnotesize
  \setlength{\tabcolsep}{2pt}
  \resizebox{\columnwidth}{!}{
  \begin{tabular}{@{}lccccc@{}}
    \toprule
    \textbf{$\lambda_{\text{tm}}$} & \textbf{Macro-F1} & \textbf{W-F1} & \textbf{Top-3} & \textbf{Cum70} & \textbf{JS} \\
    \midrule
    0.0 & .309\,{\scriptsize$\pm$.011} & .495\,{\scriptsize$\pm$.008} & \textbf{.789}\,{\scriptsize$\pm$.011} & .913\,{\scriptsize$\pm$.019} & .299\,{\scriptsize$\pm$.017} \\
    0.2 & .314\,{\scriptsize$\pm$.013} & .496\,{\scriptsize$\pm$.010} & .783\,{\scriptsize$\pm$.014} & .915\,{\scriptsize$\pm$.016} & .258\,{\scriptsize$\pm$.013} \\
    0.5 & \textbf{.319}\,{\scriptsize$\pm$.014} & .499\,{\scriptsize$\pm$.012} & .781\,{\scriptsize$\pm$.013} & .916\,{\scriptsize$\pm$.014} & .225\,{\scriptsize$\pm$.011} \\
    1.0 & .315\,{\scriptsize$\pm$.015} & .499\,{\scriptsize$\pm$.012} & .777\,{\scriptsize$\pm$.013} & \textbf{.918}\,{\scriptsize$\pm$.013} & .205\,{\scriptsize$\pm$.008} \\
    1.5 & .317\,{\scriptsize$\pm$.013} & \textbf{.501}\,{\scriptsize$\pm$.010} & .773\,{\scriptsize$\pm$.011} & .916\,{\scriptsize$\pm$.009} & \textbf{.201}\,{\scriptsize$\pm$.007} \\
    \bottomrule
  \end{tabular}
  }
  \caption{Cross-dataset validation on HOPE (15 English counselling dialogue act classes, 5-fold CV, mean $\pm$ std).}
  \label{tab:cross_dataset}
\end{table}

Results indicate generalization across languages (German to English), counselling modalities (online text-based to spoken), and category systems (60-class OnCoCo taxonomy to 15-class HOPE scheme). Macro-F1 improves from 0.309 to 0.319 (+3.2\% relative) at $\lambda_{\text{tm}}$=0.5, and JS divergence drops from 0.299 to 0.201 (33\% reduction).


\subsubsection{Label Smoothing Comparison}
\label{sec:ls_comparison}

Table~\ref{tab:ablation_ls_full} compares transition-matrix regularization against label smoothing \citep{szegedy_rethinking_2015} across all seven encoders. TM regularization consistently outperforms label smoothing on all encoders, with a mean improvement of +0.023 macro-F1 over the best label smoothing configuration. The relative gains are largest for weaker models (GELECTRa-base: +0.036, ModernGBERT-134M: +0.035), confirming that transition-based priors provide the strongest benefits when baseline performance is low.

\begin{table}[t]
\centering
\resizebox{\columnwidth}{!}{
\begin{tabular}{lccccc}
\toprule
& \multicolumn{3}{c}{\textbf{Label Smoothing}} & \multicolumn{2}{c}{\textbf{TM Reg.}} \\
\cmidrule(lr){2-4} \cmidrule(lr){5-6}
\textbf{Encoder} & $\epsilon$=0.0 & $\epsilon$=0.1 & $\epsilon$=0.2 & Best & $\Delta$ \\
\midrule
GBERT-large & .079 & .074 & .069 & \textbf{.102} & +.023 \\
GBERT-base & .073 & .081 & .071 & \textbf{.098} & +.017 \\
ModernGBERT-1B & .059 & .074 & .065 & \textbf{.096} & +.022 \\
EuroBERT-610M & .085 & .078 & .077 & \textbf{.096} & +.010 \\
EuroBERT-210M & .075 & .066 & .067 & \textbf{.095} & +.019 \\
ModernGBERT-134M & .049 & .050 & .047 & \textbf{.086} & +.035 \\
GELECTRa-base & .033 & .033 & .031 & \textbf{.070} & +.036 \\
\midrule
\textit{Mean} & .065 & .065 & .061 & \textbf{.092} & +.023 \\
\bottomrule
\end{tabular}
}
\caption{Label smoothing vs.\ TM regularization (macro-F1, 60 categories, 5-fold CV). $\Delta$ = TM best $-$ best LS.}
\label{tab:ablation_ls_full}
\end{table}

\subsubsection{Model Comparison and Effect of Regularization Strength}
\label{sec:results_model}

Figure~\ref{fig:tm_effect} visualizes the effect of transition loss weight ($\lambda_{\text{tm}}$) across architectures and metrics. Dialogue-flow metrics (Cum70, JS divergence) improve monotonically with increasing $\lambda_{\text{tm}}$, while predictive metrics (Macro-F1, Top-3) peak around $\lambda_{\text{tm}}$=1.0--1.5. The right panel shows that TM regularization provides larger effect sizes than architecture changes (BERT$\to$History).

Table~\ref{tab:encoder_results} compares performance across seven German BERT variants (110M--1B parameters), two LLMs, and baselines. A transition matrix baseline that predicts using only empirical $P(\text{next}|\text{prev})$ from training data achieves macro-F1 of 0.056, outperforming RNN models (0.003--0.008) but falling well short of fine-tuned encoders (0.070--0.102). Among fine-tuned encoders, GBERT-large achieves the highest macro-F1 (0.102), while smaller models like ModernGBERT-134M achieve competitive performance with TM regularization. The Cfg column indicates the best architecture (B=BERT, H=History) and context length for each encoder. History-augmented models dominate, with optimal context lengths of 4--12 utterances. LLMs (GPT-5-mini, gpt-oss-120b) achieve lower dialogue-flow alignment (Cum70: 0.40--0.55) despite competitive macro-F1, indicating that pretraining alone does not capture transition patterns.

\begin{table}[t]
  \centering
  \footnotesize
  \resizebox{\columnwidth}{!}{
  \begin{tabular}{@{}lclccccc@{}}
    \toprule
    \textbf{Model} & \textbf{$\lambda_{\text{tm}}$} & \textbf{Cfg} & \textbf{Macro-F1} & \textbf{W-F1} & \textbf{Top-3} & \textbf{Cum70} & \textbf{JS} \\
    \midrule
    \multicolumn{8}{l}{\textit{LLM Baselines (zero-shot, 5-fold CV)}} \\
    GPT-5-mini & -- & -- & .091\,{\scriptsize$\pm$.011} & .132\,{\scriptsize$\pm$.014} & .254\,{\scriptsize$\pm$.014} & .550\,{\scriptsize$\pm$.035} & -- \\
    gpt-oss-120b & -- & -- & .072\,{\scriptsize$\pm$.006} & .108\,{\scriptsize$\pm$.007} & .174\,{\scriptsize$\pm$.010} & .400\,{\scriptsize$\pm$.021} & -- \\
    \midrule
    \multicolumn{8}{l}{\textit{RNN Baselines (5-fold CV)}} \\
    Simple RNN & 0.5 & -- & .003\,{\scriptsize$\pm$.000} & .014\,{\scriptsize$\pm$.002} & .208\,{\scriptsize$\pm$.012} & .751\,{\scriptsize$\pm$.037} & .198\,{\scriptsize$\pm$.006} \\
    Tanaka (2019) & 1.5 & -- & .008\,{\scriptsize$\pm$.002} & .031\,{\scriptsize$\pm$.008} & .235\,{\scriptsize$\pm$.017} & .802\,{\scriptsize$\pm$.035} & .209\,{\scriptsize$\pm$.015} \\
    \midrule
    \multicolumn{8}{l}{\textit{Transition Matrix Baseline (5-fold CV)}} \\
    TM Only & -- & -- & .056\,{\scriptsize$\pm$.003} & .115\,{\scriptsize$\pm$.011} & .315 & 1.00 & -- \\
    \midrule
    \multicolumn{8}{l}{\textit{Fine-tuned Encoders (best config, 5-fold CV)}} \\
    GB-large & 0.5 & H4 & \textbf{.102\,{\scriptsize$\pm$.007}} & \textbf{.171\,{\scriptsize$\pm$.012}} & .342\,{\scriptsize$\pm$.023} & .894\,{\scriptsize$\pm$.013} & .175\,{\scriptsize$\pm$.004} \\
    GB-base & 0.5 & B8 & .098\,{\scriptsize$\pm$.010} & .158\,{\scriptsize$\pm$.015} & \textbf{.354\,{\scriptsize$\pm$.024}} & .933\,{\scriptsize$\pm$.008} & .143\,{\scriptsize$\pm$.006} \\
    MGB-1B & 0.2 & H8 & .097\,{\scriptsize$\pm$.011} & .159\,{\scriptsize$\pm$.021} & .314\,{\scriptsize$\pm$.026} & .821\,{\scriptsize$\pm$.022} & .235\,{\scriptsize$\pm$.015} \\
    EB-610M & 0.5 & H4 & .097\,{\scriptsize$\pm$.012} & .160\,{\scriptsize$\pm$.019} & .334\,{\scriptsize$\pm$.024} & .939\,{\scriptsize$\pm$.007} & .109\,{\scriptsize$\pm$.007} \\
    EB-210M & 0.5 & H12 & .095\,{\scriptsize$\pm$.015} & .158\,{\scriptsize$\pm$.020} & .331\,{\scriptsize$\pm$.027} & .937\,{\scriptsize$\pm$.011} & .125\,{\scriptsize$\pm$.004} \\
    MGB-134M & 0.5 & H4 & .086\,{\scriptsize$\pm$.007} & .147\,{\scriptsize$\pm$.009} & .358\,{\scriptsize$\pm$.021} & \textbf{.942\,{\scriptsize$\pm$.005}} & .122\,{\scriptsize$\pm$.001} \\
    GELECTRa & 1.0 & H4 & .070\,{\scriptsize$\pm$.007} & .128\,{\scriptsize$\pm$.011} & .335\,{\scriptsize$\pm$.015} & .933\,{\scriptsize$\pm$.019} & \textbf{.121\,{\scriptsize$\pm$.003}} \\
    \bottomrule
  \end{tabular}
  }
  \caption{Results by model (60 categories, 5-fold CV). Cfg = best architecture (B=BERT, H=History) + context length.}
  \label{tab:encoder_results}
\end{table}

\paragraph{Significance Testing.} Paired bootstrap tests with Benjamini-Hochberg correction confirm that TM regularization yields robust but configuration-dependent improvements, with mid-sized encoders showing the most consistent gains (Table~\ref{tab:significance_by_lambda}). History-based architectures provide benefits only for specific encoders and do not consistently outperform standard BERT (see Appendix~\ref{app:significance} for encoder-specific results).

\section{Discussion}

The results highlight several insights:

\textbf{1. Transition regularization provides consistent benefits:} The transition-matrix regularizer improves all metrics across configurations (Table~\ref{fig:tm_effect}). This dual improvement in predictive accuracy and dialogue-flow alignment reflects the fact that NDAP often admits multiple valid continuations: cross-entropy treats all non-gold predictions as equally wrong, while TM regularization provides \emph{distributional} supervision encoding plausible next acts. Moreover, the proposed regularizer is genuinely model- and architecture-agnostic: it yields consistent gains across all seven encoder variants (110M--1B parameters) and both attention-pooling and history-augmented architectures, and can be integrated as a drop-in objective without any architectural changes or sequence-level decoding.

\textbf{2. Structural priors matter more than model scale:} Across a 10$\times$ size range (110M--1B parameters), TM regularization provides larger gains than encoder choice (Table~\ref{tab:encoder_results}). ModernGBERT-1B achieves best macro-F1 at $\lambda_{\text{tm}}$=0.2 but exhibits lower dialogue-flow alignment (Cum70=.822, JS=.236) than smaller models with higher $\lambda_{\text{tm}}$---larger models may require less aggressive regularization but still benefit from structural priors.

\textbf{3. Differential benefits across encoders and architectures:} 
Encoder-level analysis (Appendix~\ref{app:significance}) reveals a clear separation. 
TM regularization acts as a general-purpose inductive bias that disproportionately benefits weaker encoders (GELECTRa-base: +0.036, ModernGBERT-134M: +0.035 macro-F1), whereas explicit history modeling provides selective gains that do not generalize across encoder families (GELECTRa: +1.93 points; GBERT-base: $-$0.16 points). 
Across all encoders, domain-specific transition priors consistently outperform uniform label smoothing (Table~\ref{tab:ablation_ls_full}).

\textbf{4. LLM comparison validates transition regularization:} In our zero-shot setting, GPT-5-mini achieves lower macro-F1 than the best fine-tuned model in Table~\ref{tab:encoder_results} and substantially lower dialogue-flow alignment (Cum70: 0.550 vs 0.894). The open-source gpt-oss-120b performs worse (macro-F1: 0.072) due to unreliable schema adherence in structured outputs, resulting in a ~17\% parse error rate. The low Cum70 for both LLMs indicates that zero-shot prompting alone does not reliably recover dialogue-flow transition patterns, supporting the use of explicit structural priors.

\textbf{5. Client simulation subset shows strong benefits:} On a 28-category client-side subset (Appendix~\ref{sec:client_subset}), TM regularization yields +18\% weighted F1 improvement (0.225$\to$0.265) confirming that the approach is effective for the client simulation use case.

\subsection{Effect of Regularization Strength}

\textbf{Response pattern:} All metrics improve with transition-loss weight, with dialogue-flow metrics (cumulative coverage, JS divergence) continuing to improve at higher $\lambda_{\text{tm}}$ values while predictive metrics peak around $\lambda_{\text{tm}}$=1.0--1.5. The increasing trend suggests the regularizer provides a stable optimization signal without creating competing objectives.

\textbf{Implicit regularization effect:} The transition-matrix regularizer also acts as an implicit regularizer against overfitting. Without it ($\lambda_{\text{tm}}$=0.0), 83\% of runs show overfitting magnitude $>$0.5. Positive $\lambda_{\text{tm}}$ reduces this: at $\lambda_{\text{tm}}$=1.5, overfitting magnitude falls to 0.20 with no runs exceeding 0.5.

\textbf{Practical recommendation:} For practitioners, a $\lambda_{\text{tm}} \in [0.5, 1.0]$ as a default is recommended. Across all configurations, $\lambda_{\text{tm}}$=0.5 most frequently achieves optimal macro-F1, while $\lambda_{\text{tm}}$=1.0 performs within 5\% of optimal in over 70\% of configurations, making it a robust choice when extensive tuning is not feasible. This pattern also holds for HOPE. 

\section{Conclusion}

This paper proposed a transition-matrix KL regularizer for NDAP that incorporates empirical dialogue-flow structure into the training objective. Across experiments on a fine-grained German counselling taxonomy and cross-dataset transfer to other dialogue corpora, the regularizer consistently improves both predictive performance and alignment with observed dialogue-flow dynamics. Additionally, we presented history-augmented architectures that leverage broader dialogue context through multi-head attention.

A natural next step is integrating NDAP with LLM control, using predicted categories to condition prompts or guide sampling strategies for client simulation. This could also connect naturally to automated feedback pipelines in counselor-training \citep{rudolph_automated_2024}. Recent adversarial evaluations of LLM-based virtual clients suggest that maintaining role consistency remains brittle under prompt attacks \citep{rudolph_evaluating_2026}, making structured priors like transition-matrix regularization a plausible complementary control signal. Additionally, fine-tuning decoder-only architectures on NDAP could leverage their autoregressive nature to better model dialogue sequences while potentially benefiting from transition-matrix regularization.

\section*{Limitations}

The primary dataset consists of 76 conversations with approximately 5,500 utterances---a relatively small corpus for training neural models. While cross-validation and cross-dataset experiments provide evidence of robustness, the limited size restricts conclusions about generalization to larger-scale deployments or substantially different counselling contexts.

The conversations are role-play sessions conducted by social science students, not recordings of real clinical counselling. While participants followed structured scenarios designed to reflect authentic counselling dynamics, student role-plays may lack the pragmatic complexity, emotional depth, and unpredictability of genuine therapeutic interactions. Models trained on this data may not fully capture the nuances present in real-world counselling.

Annotations were produced by trained students and subsequently reviewed by a domain expert in a sequential process. This workflow precludes formal inter-rater agreement metrics, which are standard for validating annotation reliability. Although expert review ensures quality control, it cannot replace an independent double-annotation study with quantified agreement and therefore limits assessment of annotation consistency and reproducibility.

The transition matrix loss assumes transitions are stable across the corpus; in applications with significant domain shift or concept drift, alternative regularization approaches may be needed. Our evaluation uses gold dialogue act labels for the conversation history; although this is common practice in NDAP, these markings would have to be predicted in fully autonomous operations, which could potentially lead to chain errors. Our LLM comparison is also limited to two zero-shot baselines; stronger prompted or fine-tuned decoder models remain future work. End-to-end evaluation combining NDAP with LLM-based response generation remains unexplored.


\section*{Ethical considerations}

The counselling dialogue data used in this study were collected from structured role-play sessions conducted by social science students in a university course setting. Participants acted in predefined client and counselor roles; no real clients, patients, or personal therapeutic information were involved. All participants provided informed consent for research use of the data, and all conversations were anonymized prior to analysis. The dialogues were annotated by paid social science students with prior training in counselling concepts. Annotation included both dialogue act labeling and anonymization of the conversations. All annotations were subsequently reviewed by a domain expert to ensure consistency and quality.

We acknowledge that dialogue act prediction technology for counselling contexts could potentially be misused. Recent conceptual work distinguishes autonomous counselor bots, AI training simulators, and counselor-facing augmentation tools as ethically distinct implementation approaches \citep{steigerwald_ai_2026}. Our work is positioned primarily in the training-simulator setting and secondarily as a component for counselor-facing support, not as an autonomous counselling agent. The intended application, controllable client simulation for counselor training, serves an educational purpose that may improve the quality of counseling services. The data set released contains only simulated conversations and poses minimal privacy risk.

\section*{Acknowledgments}

Generative AI tools (ChatGPT and Claude) were used throughout the article for coding assistance, to identify relevant literature, LaTeX figure and table preparation, and to improve clarity and style of writing. They were not used to generate scientific claims or experimental results. All content was reviewed, verified, and finalized by the authors.

\bibliography{references}

\appendix

\section{Model Architecture Details}

\subsection{Hyperparameter Settings}

\begin{table}[h]
  \centering
  \small
  \begin{tabular}{ll}
    \toprule
    \textbf{Hyperparameter} & \textbf{Value} \\
    \midrule
    \multicolumn{2}{l}{\textit{Training Configuration}} \\
    Epochs & 10 \\
    Batch size & 64 \\
    Learning rate scheduler & Cosine \\
    Gradient clipping & 1.0 \\
    Mixed precision (AMP) & Enabled \\
    \midrule
    \multicolumn{2}{l}{\textit{Early Stopping}} \\
    Patience & 3 epochs \\
    Min delta & 0.001 \\
    Monitor metric & Validation macro-F1 \\
    \midrule
    \multicolumn{2}{l}{\textit{Model Architecture}} \\
    Context utterances & \{1, 4, 8, 12\} \\
    History length & 10 categories \\
    Attention heads & 8 \\
    Transformer dropout & 0.2 \\
    \midrule
    \multicolumn{2}{l}{\textit{Regularization Grid}} \\
    Numerical stability constant $\epsilon_{\text{num}}$ & $10^{-8}$ (fixed) \\
    $\lambda_{\text{tm}}$ & \{0.0, 0.2, 0.5, 1.0, 1.5\} \\
    Label smoothing $\epsilon$ & \{0.0, 0.1, 0.2\} \\
    \midrule
    \multicolumn{2}{l}{\textit{Cross-Validation}} \\
    Number of folds & 5 \\
    Split strategy & Conversation-level \\
    Random seed & 42 \\
    \bottomrule
  \end{tabular}
  \caption{Hyperparameter settings. Context utterances and regularization weights were varied in grid search; other values were fixed across all experiments.}
\end{table}

\subsection{Ablation Study}

Ablation studies are conducted to validate the design choices: (1) evaluating on a client simulation subset, (2) comparing transition-matrix regularization against label smoothing, (3) evaluating the effect of synthetic data augmentation, (4) exploring alternative KL formulations, and (5) investigating hierarchical label embeddings.

\subsubsection{Client Simulation Subset}
\label{sec:client_subset}

For client simulation applications, predicting client-side dialogue acts is the primary goal. This subset directly matches the virtual-client training setting explored in VirCo \citep{rudolph_ai-based_2024}. TM regularization is evaluated on a subset containing only client-side categories, filtering for counselor$\to$client and client$\to$client transitions.

\paragraph{Dataset.} The client simulation subset contains 2,176 instances across 28 client-side categories. A train/test split of 1,604/572 utterances with conversation-level partitioning is used. The following ablation studies (Sections~\ref{sec:ablation}--\ref{sec:ablation_synth}) are conducted on this subset to provide additional insights into design choices.

\paragraph{Effect of Transition Regularization.} Table~\ref{tab:client_tm} shows the effect of varying $\lambda_{\text{tm}}$ on the client simulation subset.

\begin{table}[h]
\centering
\footnotesize
\begin{tabular}{lcccc}
\toprule
$\lambda_{\text{tm}}$ & W-F1 & Top-3 & Trans. & Cum70 \\
\midrule
0.0 & .225 & .451 & .599 & .801 \\
0.2 & .255 & .495 & .743 & .894 \\
0.5 & \textbf{.265} & \textbf{.505} & .799 & .925 \\
1.0 & .251 & .497 & .838 & .955 \\
1.5 & .253 & .494 & \textbf{.875} & \textbf{.961} \\
\bottomrule
\end{tabular}
\caption{Effect of $\lambda_{\text{tm}}$ on client simulation subset (28 categories). Trans.=transition validity. Cum70=cumulative accuracy at 70\%.}
\label{tab:client_tm}
\end{table}

TM regularization yields +18\% weighted F1 improvement (0.225$\to$0.265 at $\lambda_{\text{tm}}$=0.5), stronger than the +17\% improvement on the full 60-category task. Transition validity improves by +46\% (0.599$\to$0.875 at $\lambda_{\text{tm}}$=1.5).

\paragraph{Encoder Comparison.} Table~\ref{tab:client_encoders} compares encoder performance on the client simulation subset.

\begin{table}[h]
\centering
\footnotesize
\begin{tabular}{lccc}
\toprule
\textbf{Encoder} & \textbf{W-F1} & \textbf{Top-3} & \textbf{Params} \\
\midrule
GBERT-large & \textbf{.292} & .506 & 336M \\
ModernGBERT-1B & .287 & .507 & 1B \\
GBERT-base & .282 & \textbf{.518} & 110M \\
EuroBERT-610M & .281 & .492 & 610M \\
EuroBERT-210M & .274 & .513 & 210M \\
ModernGBERT-134M & .271 & .538 & 134M \\
GELECTRa-base & .252 & .517 & 110M \\
\bottomrule
\end{tabular}
\caption{Encoder comparison on client simulation subset (28 categories, best $\lambda_{\text{tm}}$ per encoder). Model size (110M--1B) shows no consistent correlation with performance.}
\label{tab:client_encoders}
\end{table}

Best weighted F1 (0.292) is achieved with GBERT-large at $\lambda_{\text{tm}}$=0.2. Notably, the absolute F1 scores on the 28-category subset are substantially higher than on the 60-category task (0.292 vs 0.164), reflecting the reduced complexity of the classification problem. However, the relative improvement from TM regularization remains consistent, confirming that dialogue-flow priors are valuable across task granularities.

\subsubsection{Transition-Matrix vs. Label Smoothing}
\label{sec:ablation}

Table~\ref{tab:ablation_ls} compares our transition-matrix regularization against label smoothing \citep{szegedy_rethinking_2015} with $\epsilon \in \{0.0, 0.1, 0.2\}$ on the client simulation subset. While label smoothing provides modest gains for some architectures, transition-matrix regularization consistently outperforms across most encoder variants. Note that this label-smoothing parameter is distinct from the fixed numerical-stability constant $\epsilon_{\text{num}} = 10^{-8}$ used for transition-matrix computations.

\begin{table}[h]
\centering
\resizebox{\columnwidth}{!}{
\begin{tabular}{lccccc}
\toprule
& \multicolumn{3}{c}{\textbf{Label Smoothing}} & \multicolumn{2}{c}{\textbf{TM Reg.}} \\
\cmidrule(lr){2-4} \cmidrule(lr){5-6}
\textbf{Encoder} & $\epsilon$=0.0 & $\epsilon$=0.1 & $\epsilon$=0.2 & Best & $\Delta$ \\
\midrule
GBERT-large & .244 & .268 & .255 & \textbf{.292} & +.024 \\
GBERT-base & .253 & .263 & .260 & \textbf{.282} & +.019 \\
ModernGBERT-134M & .256 & .251 & \textbf{.277} & .271 & --.006 \\
GELECTRa-base & .206 & .197 & .208 & \textbf{.253} & +.044 \\
EuroBERT-210M & .243 & .223 & .252 & \textbf{.275} & +.023 \\
EuroBERT-610M & .229 & .225 & .241 & \textbf{.281} & +.041 \\
\midrule
\textit{Mean} & .239 & .238 & .249 & \textbf{.276} & +.024 \\
\bottomrule
\end{tabular}
}
\caption{Label smoothing vs.\ TM regularization (weighted F1). TM outperforms on 5 of 6 encoders.}
\label{tab:ablation_ls}
\end{table}

The mean improvement of TM regularization over the best label smoothing configuration ($\epsilon$=0.2) is +0.024 F1 points. Notably, the gains are largest for architectures with lower baseline performance (GElectra-base: +4.4\%, EuroBERT-610M: +4.1\%), suggesting that domain-specific structural priors are particularly valuable when model capacity is limited. Table~\ref{tab:ablation_ls_full} in the main paper shows that this pattern holds even more strongly on the full 60-category task.

\subsubsection{Synthetic Data Augmentation}
\label{sec:ablation_synth}

Given the substantial class imbalance in the client simulation subset (Gini coefficient 0.51), LLM-based synthetic data augmentation was explored. A two-phase strategy was employed: (1) prompt-based augmentation using GPT-5-mini to balance minority classes up to 100 examples each, and (2) persona-based generation using expert-designed client profiles to introduce lexical and stylistic diversity. The augmented training set contains 8,487 instances (81\% synthetic), reducing the Gini coefficient from 0.51 to 0.06.

Table~\ref{tab:ablation_synth} compares BERT-based models trained on real data only versus real+synthetic data on the client simulation subset.

\begin{table}[h]
\centering
\small
\begin{tabular}{lccc}
\toprule
\textbf{Encoder} & \textbf{Real} & \textbf{+Synth.} & $\Delta$ \\
\midrule
GBERT-large & \textbf{.292} & .281 & --.011 \\
GBERT-base & \textbf{.282} & .264 & --.018 \\
ModernGBERT-134M & .271 & \textbf{.283} & +.011 \\
GELECTRa-base & .253 & \textbf{.257} & +.004 \\
EuroBERT-210M & .275 & \textbf{.283} & +.008 \\
EuroBERT-610M & \textbf{.281} & .276 & --.006 \\
\midrule
\textit{Mean} & \textbf{.276} & .274 & --.002 \\
\bottomrule
\end{tabular}
\caption{Real vs.\ real+synthetic training (weighted F1). Synthetic augmentation provides no consistent benefit for BERT-based models.}
\label{tab:ablation_synth}
\end{table}

Results are mixed: synthetic data slightly improves smaller models (ModernGBERT-134M, EuroBERT-210M) but degrades larger models (GBERT-large, GBERT-base). The overall mean shows no significant difference ($p > 0.17$, independent t-test), suggesting that pretrained transformers are sufficiently data-efficient for this task.

In contrast, RNN baselines benefited substantially from synthetic augmentation: Simple RNN achieved F1=0.234 vs 0.097 on real data only. This suggests that pretrained transformers are sufficiently data-efficient for fine-grained dialogue act prediction, while RNNs require substantially more examples to converge. For this task, domain-specific structural priors (like transition matrices) appear more valuable than quantity-based augmentation when using pretrained encoders.

\subsubsection{Cross-Dataset Validation on SWDA and Alternative KL Formulations}
\label{sec:ablation_kl}

To test generalization beyond counselling and investigate the effect of skewed transition distributions, we evaluated TM regularization on the Switchboard Dialogue Act Corpus (SWDA; \citealp{Shriberg-etal:1998, stolcke_dialogue_2000}), a benchmark of English telephone conversations filtered to 37 classes (excluding rare categories with fewer than 50 occurrences). Table~\ref{tab:swda_cross_dataset} shows results using XLM-RoBERTa.

\begin{table}[h]
  \centering
  \footnotesize
  \begin{tabular}{@{}lcccc@{}}
    \toprule
    \textbf{$\lambda_{\text{tm}}$} & \textbf{F1} & \textbf{Top-3} & \textbf{Cum70} & \textbf{JS} \\
    \midrule
    0.0 & \textbf{.169} & .765 & .901 & .150 \\
    0.5 & .162 & .767 & .929 & .073 \\
    1.5 & .144 & \textbf{.771} & \textbf{.986} & \textbf{.023} \\
    \bottomrule
  \end{tabular}
  \caption{SWDA 37-class results. Dialogue-flow metrics (Cum70, JS) improve substantially but macro-F1 decreases with TM regularization.}
  \label{tab:swda_cross_dataset}
\end{table}

Unlike counselling datasets (OnCoCo, HOPE), TM regularization on SWDA improves dialogue-flow alignment (JS: 0.150$\to$0.023, Cum70: 0.901$\to$0.986) but \emph{decreases} macro-F1. This divergence reflects SWDA's highly skewed transition distribution: analysis of the empirical transition matrix reveals that 69\% of source classes have a single dominant successor---the ``Statement-non-opinion'' (sd) category, which accounts for 34\% of all transitions. The regularizer thus pushes predictions toward this dominant class, improving flow alignment but suppressing minority class predictions.

To address this skewness, alternative formulations of the transition-matrix loss were explored: (1) reverse KL divergence, which is more permissive for confident predictions, and (2) entropy-weighted KL, which down-weights samples from high-entropy source categories where multiple successors are plausible. Neither variant improved over standard forward KL (Table~\ref{tab:ablation_kl}), suggesting that the macro-F1 degradation on SWDA stems from the dataset's inherent transition structure rather than the KL formulation.

\begin{table}[h]
\centering
\small
\begin{tabular}{lcc}
\toprule
\textbf{KL Variant} & \textbf{Accuracy} & \textbf{Macro-F1} \\
\midrule
Forward KL (baseline) & \textbf{0.502} & \textbf{0.150} \\
Reverse KL & 0.500 & 0.149 \\
Entropy-weighted & 0.496 & 0.146 \\
Entropy + Reverse & 0.495 & 0.146 \\
\bottomrule
\end{tabular}
\caption{Alternative KL formulations on SWDA 37-class ($\lambda_{\text{tm}}$=0.5). Forward KL performs best.}
\label{tab:ablation_kl}
\end{table}

This finding suggests that TM regularization is most effective when transition patterns are more balanced, as in counselling domains where dialogue follows structured phases (problem exploration $\to$ intervention $\to$ resolution) rather than converging on a single dominant act type.

\subsubsection{Context Window Size}
\label{sec:ablation_context}

The number of preceding utterances used as context is varied (1, 4, 8, 12). Table~\ref{tab:context_ablation} shows macro-F1 across context lengths and TM weights. Performance peaks at 4 utterances with $\lambda \geq 0.5$ (macro-F1=0.080). Longer context windows provide diminishing returns, suggesting that recent dialogue history is most informative for NDAP.

\begin{table}[h]
\centering
\footnotesize
\resizebox{\columnwidth}{!}{
\begin{tabular}{lccccc}
\toprule
& \multicolumn{5}{c}{\textbf{TM Weight ($\lambda_{\text{tm}}$)}} \\
\cmidrule(lr){2-6}
\textbf{Ctx} & 0.0 & 0.2 & 0.5 & 1.0 & 1.5 \\
\midrule
1 & .065\,{\scriptsize$\pm$.012} & .069\,{\scriptsize$\pm$.012} & .072\,{\scriptsize$\pm$.010} & .071\,{\scriptsize$\pm$.011} & .073\,{\scriptsize$\pm$.009} \\
4 & .070\,{\scriptsize$\pm$.019} & .073\,{\scriptsize$\pm$.020} & \textbf{.080\,{\scriptsize$\pm$.013}} & \textbf{.080\,{\scriptsize$\pm$.014}} & \textbf{.080\,{\scriptsize$\pm$.011}} \\
8 & .066\,{\scriptsize$\pm$.018} & .076\,{\scriptsize$\pm$.015} & .074\,{\scriptsize$\pm$.020} & .079\,{\scriptsize$\pm$.013} & .079\,{\scriptsize$\pm$.010} \\
12 & .065\,{\scriptsize$\pm$.013} & .070\,{\scriptsize$\pm$.019} & .075\,{\scriptsize$\pm$.018} & .077\,{\scriptsize$\pm$.014} & .077\,{\scriptsize$\pm$.012} \\
\bottomrule
\end{tabular}
}
\caption{Macro-F1 by context window size and TM weight (mean $\pm$ std, 5-fold CV). Best at 4 utterances with $\lambda \geq 0.5$.}
\label{tab:context_ablation}
\end{table}

\subsubsection{Hierarchical Label Embeddings}
\label{sec:ablation_hierarchy}

The OnCoCo taxonomy organizes 60 leaf categories into a five-level hierarchy. Experiments explored whether explicitly encoding this structure could improve predictions by embedding each hierarchy level ($K_1$--$K_5$) independently and integrating them with the conversation context via cross-attention. This approach is inspired by hierarchical text classification methods \citep{kowsari_hdltex_2017}. However, experiments showed only marginal improvements over the base architecture (+0.5\% weighted F1), insufficient to justify the additional complexity. The hypothesis is that the pretrained encoder already captures sufficient semantic structure, and that the transition-matrix regularizer---which implicitly encodes label relationships through observed co-occurrence patterns---provides a more effective inductive bias than explicit hierarchy embeddings.

\subsubsection{LLM Baseline Prompt Template}
\label{sec:llm_prompt}

The following prompt template is used for GPT-5-mini zero-shot evaluation. Category descriptions and conversation history are inserted at the indicated placeholders.

\begin{small}
\begin{verbatim}
You are an expert in dialogue act classification
for German online counseling.

## Task
Predict the dialogue act category of the NEXT
utterance in this counseling conversation.

## Categories (60 total)
[CATEGORY_CODE]: [DESCRIPTION]
... (all 60 categories with descriptions)

## Conversation History (last 12 turns)
[SPEAKER] ([CATEGORY_CODE] | [DESCRIPTION]): [TEXT]
... (conversation turns with speaker, category, text)

## Output Format (JSON)
Return your top 3 predictions:
{
  "predictions": [
    {"category": "CODE", "confidence": 0.6},
    {"category": "CODE", "confidence": 0.25},
    {"category": "CODE", "confidence": 0.15}
  ]
}
\end{verbatim}
\end{small}

\section{Significance Tests by Encoder}
\label{app:significance}

We report paired bootstrap significance tests (10,000 iterations) with Benjamini-Hochberg FDR correction for multiple comparisons. Table~\ref{tab:significance_by_lambda} summarizes results across regularization strengths, demonstrating consistent positive effects at all $\lambda_{\text{tm}}$ values tested. Tables~\ref{tab:tm_effect_by_encoder} and~\ref{tab:arch_effect_by_encoder} show encoder-specific results.

\begin{table}[h]
\centering
\footnotesize
\resizebox{\columnwidth}{!}{
\begin{tabular}{@{}rrrrrrrr@{}}
\toprule
\textbf{$\lambda_{\text{tm}}$} & \textbf{Tests} & \textbf{Pos.} & \textbf{\%Pos.} & \textbf{Sig.} & \textbf{\%Sig.} & \textbf{Mean $\Delta$} & \textbf{Med. $\Delta$} \\
\midrule
    0.2 & 42 & 36 & 86\% & 9 & 21\% & +0.0061 & +0.0050 \\
    0.5 & 44 & 37 & 84\% & 25 & 57\% & +0.0069 & +0.0063 \\
    1.0 & 45 & 40 & 89\% & 22 & 49\% & +0.0089 & +0.0091 \\
    1.5 & 42 & 33 & 79\% & 19 & 45\% & +0.0075 & +0.0070 \\
\bottomrule
\end{tabular}
}
\caption{Summary of TM regularization significance tests by $\lambda_{\text{tm}}$ value. Tests: paired comparisons (encoder $\times$ architecture $\times$ context). Pos.: positive effect over $\lambda_{\text{tm}}$=0. Sig.: significant after FDR correction ($\alpha$=0.05).}
\label{tab:significance_by_lambda}
\end{table}

\begin{table}[h]
\centering
\footnotesize
\begin{tabular}{@{}lcc@{}}
\toprule
\textbf{Encoder} & \textbf{$\Delta$ Macro-F1} & \textbf{Sig. (FDR$<$0.05)} \\
\midrule
ModernGBERT-134M & +0.93\% & 5/8 (62\%) \\
EuroBERT-610M & +0.80\% & 4/5 (80\%) \\
ModernGBERT-1B & +0.72\% & 3/7 (43\%) \\
GBERT-base & +0.70\% & 5/6 (83\%) \\
GELECTRa-base & +0.68\% & 4/8 (50\%) \\
GBERT-large & +0.47\% & 3/7 (43\%) \\
EuroBERT-210M & +0.38\% & 1/3 (33\%) \\
\midrule
\textbf{Overall} & \textbf{+0.69\%} & \textbf{25/44 (57\%)} \\
\bottomrule
\end{tabular}
\caption{TM effect ($\lambda_{\text{tm}}$=0 vs 0.5) by encoder. Mid-sized encoders show highest significance rates.}
\label{tab:tm_effect_by_encoder}
\end{table}

\begin{table}[h]
\centering
\footnotesize
\begin{tabular}{@{}lcc@{}}
\toprule
\textbf{Encoder} & \textbf{$\Delta$ Macro-F1} & \textbf{Sig. (FDR$<$0.05)} \\
\midrule
GELECTRa-base & +1.93\% & 4/4 (100\%) \\
ModernGBERT-134M & +0.93\% & 2/4 (50\%) \\
EuroBERT-210M & +0.75\% & 2/3 (67\%) \\
EuroBERT-610M & +0.64\% & 2/4 (50\%) \\
ModernGBERT-1B & +0.50\% & 1/4 (25\%) \\
GBERT-large & $-$0.08\% & 1/3 (33\%) \\
GBERT-base & $-$0.16\% & 1/4 (25\%) \\
\midrule
\textbf{Overall} & \textbf{+0.67\%} & \textbf{13/26 (50\%)} \\
\bottomrule
\end{tabular}
\caption{Architecture effect (BERT$\to$History, $\lambda_{\text{tm}}$=0.5) by encoder. Benefits weaker encoders only.}
\label{tab:arch_effect_by_encoder}
\end{table}

\end{document}

%% file: figures/contribution_figure.tex
\begin{center}
\resizebox{\columnwidth}{!}{
\begin{tikzpicture}[
    node distance=0.3cm,
    box/.style={rectangle, draw, rounded corners, minimum height=0.5cm, align=center, font=\scriptsize},
    label/.style={font=\scriptsize},
    smalllabel/.style={font=\tiny},
    bar/.style={rectangle, minimum height=0.25cm},
]

\begin{scope}[local bounding box=left]
    \node[font=\scriptsize\bfseries] (lefttitle) at (0, 2.8) {(a) Standard Training};

    \node[box, fill=gray!30, text width=1.4cm, align=center] (da1) at (-0.85, 2.1) {
        \textbf{Prev. act:}\\
        \textit{Concern Inq.}
    };

    \node[box, fill=yellow!20, text width=1.5cm, align=center] (text1) at (1, 2.1) {
        ``How can\\I help?''
    };

    \node[box, fill=blue!15] (model1) at (0, 1.3) {Model};
    \draw[->, >=latex, dashed] (da1.south) |- (model1.west);
    \draw[->, >=latex] (text1.south) |- (model1.east);

    \node[smalllabel, anchor=west] at (-1.8, 0.95) {Predicted $\hat{p}$:};

    \fill[blue!60] (-1.8, 0.6) rectangle (-0.3, 0.8);  
    \node[smalllabel, anchor=west] at (-0.2, 0.7) {Problem Desc.};

    \fill[blue!60] (-1.8, 0.25) rectangle (-1.0, 0.45);  
    \node[smalllabel, anchor=west] at (-0.2, 0.35) {Emotional Disc.};

    \fill[blue!60] (-1.8, -0.1) rectangle (-1.4, 0.1);  
    \node[smalllabel, anchor=west] at (-0.2, 0.0) {Greeting};

    \node[smalllabel, anchor=west] at (-1.8, -0.5) {Gold label:};
    \fill[green!60] (-1.8, -0.85) rectangle (-0.3, -0.65);
    \node[smalllabel, anchor=west] at (-0.2, -0.75) {Problem Desc. \textcolor{green!50!black}{\checkmark}};

    \node[smalllabel, anchor=west, text=red!70!black] at (-0.2, -1.1) {Emotional Disc. \texttimes};
    \node[smalllabel, anchor=west, text=red!70!black] at (-0.2, -1.4) {Greeting \texttimes};

    \node[box, fill=orange!20] (loss1) at (0, -2.0) {$\mathcal{L} = \text{CE}(\hat{p}, y_{\text{gold}})$};

    \node[smalllabel, text=red!70!black, text width=2.8cm, align=center] at (0, -2.6) {Penalizes plausible alternatives};
\end{scope}

\draw[gray, dashed] (2.2, 3.0) -- (2.2, -3.0);

\begin{scope}[xshift=4.4cm, local bounding box=right]
    \node[font=\scriptsize\bfseries] (righttitle) at (0, 2.8) {(b) Proposed Training};

    \node[box, fill=gray!30, text width=1.4cm, align=center] (da2) at (-0.85, 2.1) {
        \textbf{Prev. act:}\\
        \textit{Concern Inq.}
    };

    \node[box, fill=yellow!20, text width=1.5cm, align=center] (text2) at (1, 2.1) {
        ``How can\\I help?''
    };

    \node[box, fill=blue!15] (model2) at (0, 1.3) {Model};
    \draw[->, >=latex, dashed] (da2.south) |- (model2.west);
    \draw[->, >=latex] (text2.south) |- (model2.east);

    \node[smalllabel, anchor=west] at (-1.8, 0.95) {Predicted $\hat{p}$:};

    \fill[blue!60] (-1.8, 0.6) rectangle (-0.3, 0.8);
    \node[smalllabel, anchor=west] at (-0.2, 0.7) {Problem Desc.};

    \fill[blue!60] (-1.8, 0.25) rectangle (-1.0, 0.45);
    \node[smalllabel, anchor=west] at (-0.2, 0.35) {Emotional Disc.};

    \fill[blue!60] (-1.8, -0.1) rectangle (-1.4, 0.1);
    \node[smalllabel, anchor=west] at (-0.2, 0.0) {Greeting};

    \node[smalllabel, anchor=west] at (-1.8, -0.5) {Transition target $\mathbf{t}$:};
    \fill[green!50] (-1.8, -0.85) rectangle (-0.5, -0.65);
    \node[smalllabel, anchor=west] at (-0.2, -0.75) {Problem Desc.};

    \fill[green!50] (-1.8, -1.2) rectangle (-1.1, -1.0);
    \node[smalllabel, anchor=west] at (-0.2, -1.1) {Emotional Disc.};

    \fill[green!50] (-1.8, -1.55) rectangle (-1.5, -1.35);
    \node[smalllabel, anchor=west] at (-0.2, -1.45) {Greeting};

    \node[box, fill=green!20] (loss2) at (0, -2.1) {$\mathcal{L} = \text{CE} + \lambda \cdot \text{KL}(\hat{p} \| \mathbf{t})$};

    \node[smalllabel, text=green!50!black, text width=2.8cm, align=center] at (0, -2.7) {Aligns with valid transitions};
\end{scope}

\end{tikzpicture}
}
\captionsetup{hypcap=false}
\captionof{figure}{Comparison of standard NDAP training (a) vs.\ transition-matrix regularized training (b). Standard cross-entropy penalizes all non-gold predictions equally, even when multiple next acts are plausible. The proposed regularizer aligns predictions with empirical transition patterns from the corpus.}
\label{fig:contribution}
\end{center}

%% file: history_model_fig.tex
\begin{figure}[t]
\centering
\resizebox{0.95\columnwidth}{!}{
\begin{tikzpicture}[
    node distance=0.4cm and 0.5cm,
    box/.style={rectangle, draw, rounded corners, minimum height=0.6cm,
                align=center, font=\scriptsize},
    smallbox/.style={rectangle, draw, rounded corners, minimum height=0.5cm,
                minimum width=1.2cm, align=center, font=\footnotesize},
    attn/.style={rectangle, draw, rounded corners, fill=blue!8,
                minimum height=0.5cm, align=center, font=\footnotesize},
    encoder/.style={rectangle, draw, rounded corners, fill=green!8,
                minimum height=0.55cm, align=center, font=\scriptsize},
    histbox/.style={rectangle, draw, rounded corners, fill=orange!8,
                minimum height=0.5cm, align=center, font=\footnotesize},
    label/.style={font=\footnotesize, text=gray},
    arrow/.style={->, >=latex, line width=0.5pt},
    dasharrow/.style={->, >=latex, dashed, line width=0.5pt, gray},
]

\node[label, font=\footnotesize\bfseries] (baselabel) {Base (BERT)};

\node[box, below=0.3cm of baselabel, text width=2.2cm, font=\footnotesize] (input1) {Utterances $u_1, \ldots, u_n$};

\node[encoder, below=0.4cm of input1, text width=2.2cm, font=\footnotesize] (bert1) {BERT Encoder};

\node[attn, below=0.4cm of bert1, text width=2.2cm, font=\footnotesize] (pool1) {Attention Pooling};

\node[box, below=0.4cm of pool1, text width=1.8cm, font=\footnotesize] (class1) {Linear Layer};

\node[box, below=0.4cm of class1, text width=1.8cm, fill=yellow!10, font=\footnotesize] (out1) {Next DA: $\hat{c}_t$};

\draw[arrow] (input1) -- (bert1);
\draw[arrow] (bert1) -- (pool1);
\draw[arrow] (pool1) -- (class1);
\draw[arrow] (class1) -- (out1);

\node[label, right=2.5cm of baselabel, font=\footnotesize\bfseries] (histlabel) {History-Augmented};

\node[box, below=0.3cm of histlabel, text width=2.2cm, font=\footnotesize] (input2) {Utterances $u_1, \ldots, u_n$};

\node[encoder, below=0.4cm of input2, text width=2.2cm, font=\footnotesize] (bert2) {BERT Encoder};

\node[attn, below=0.4cm of bert2, text width=2.6cm] (mhsa) {Multi-Head Self-Attn};

\node[attn, below=0.4cm of mhsa, text width=2.6cm, font=\footnotesize] (cuattn) {Cross-Utt. Attn (8-head)};

\node[box, below=0.4cm of cuattn, text width=1.6cm, font=\footnotesize] (ctx) {Context $\mathbf{c}_t$};

\node[histbox, right=1.0cm of input2, text width=1.8cm, font=\footnotesize] (histin) {History $c_{t-k:t-1}$};
\node[histbox, below=0.35cm of histin, text width=1.8cm, font=\footnotesize] (embed) {Embed + Pos};
\node[attn, below=0.35cm of embed, text width=1.8cm, fill=orange!8, font=\footnotesize] (histattn) {Hist. Attn (8-head)};

\node[attn, below=0.4cm of ctx, text width=2.6cm, font=\footnotesize] (crossattn) {Text-Cat. Cross-Attn};

\node[box, below=0.4cm of crossattn, text width=1.8cm, font=\footnotesize] (class2) {Linear Layer};

\node[box, below=0.4cm of class2, text width=1.8cm, fill=yellow!10, font=\footnotesize] (out2) {Next DA: $\hat{c}_t$};

\draw[arrow] (input2) -- (bert2);
\draw[arrow] (bert2) -- (mhsa);
\draw[arrow] (mhsa) -- (cuattn);
\draw[arrow] (cuattn) -- (ctx);
\draw[arrow] (ctx) -- (crossattn);
\draw[arrow] (crossattn) -- (class2);
\draw[arrow] (class2) -- (out2);

\draw[arrow] (histin) -- (embed);
\draw[arrow] (embed) -- (histattn);
\draw[arrow] (histattn.south) |- (crossattn.east);

\begin{scope}[on background layer]
    \node[draw=gray!50, dashed, rounded corners, inner sep=0.12cm,
          fit=(histin)(embed)(histattn)] {};
\end{scope}

\end{tikzpicture}
}
\caption{Model architectures for NDAP: BERT + attention pooling \textbf{(left)} and a history-augmented variant that incorporates conversational context and previous dialogue-act labels \textbf{(right)}.}
\label{fig:history_model}
\end{figure}